\documentclass{article}

\PassOptionsToPackage{numbers,sort&compress}{natbib}
\usepackage[eandd,final]{neurips_2026}

\usepackage[utf8]{inputenc}
\usepackage[T1]{fontenc}
\usepackage{hyperref}
\usepackage{url}
\usepackage{microtype}
\usepackage{graphicx}
\usepackage{subfigure}
\usepackage{booktabs}
\usepackage{placeins}
\usepackage{float}
\usepackage{xcolor}
\usepackage{nicefrac}
\usepackage{amsmath}
\usepackage{amssymb}
\usepackage{amsfonts}
\usepackage{mathtools}
\usepackage{amsthm}
\usepackage{subcaption}
\usepackage[capitalize,noabbrev]{cleveref}
\usepackage[textsize=tiny]{todonotes}

\theoremstyle{plain}

\theoremstyle{definition}

\theoremstyle{remark}

\title{Discrete Diffusion Playground: 

A 2D Benchmark for Discrete Generative Models}

\author{%
  Justin Shao\textsuperscript{*}\\
  Carnegie Mellon University\\
  \texttt{jshao2@andrew.cmu.edu}
  \And
  Sihyun Park\textsuperscript{*}\\
  Carnegie Mellon University\\
  \texttt{sihyunpa@andrew.cmu.edu}
  \AND
  David Koes\\
  University of Pittsburgh\\
  \texttt{dkoes@pitt.edu}
  \And
  Maria Chikina\\
  University of Pittsburgh\\
  \texttt{mchikina@pitt.edu}
}

\begin{document}

\maketitle

\begingroup
\renewcommand{\thefootnote}{*}
\footnotetext{Equal contribution.}
\endgroup

\begin{abstract}
Discrete diffusion is a promising framework for categorical generative modeling, but current evaluation protocols often fail to reveal whether models have actually learned the target distribution. We introduce the Discrete Diffusion Playground, a diagnostic benchmark that converts known 2D continuous distributions into binary token sequences through an exact discretized bijection. This creates a rare setting in which models are trained on discrete sequences, but generated samples can be decoded back into 2D and compared directly against a known ground-truth distribution. The benchmark spans smooth Gaussian mixtures and tilted checkerboards with sharp boundaries, disconnected support, and coordinate-dependent structure, allowing it to expose failures that are invisible in standard proxy metrics. It supports exact grid divergences, Wasserstein distances, density visualizations, and nonparametric two-sample tests, providing both quantitative evaluation and interpretable failure diagnosis. Using this testbed, we show that masked diffusion models can learn meaningful adaptive reveal orders but remain less stable than autoregressive baselines on smooth targets and largely fail on tilted checkerboards, where autoregressive models preserve the target geometry. Control experiments with reversed and interleaved token orderings show that autoregressive performance is not explained by favorable bit order alone. The Playground provides a simple but stringent benchmark for stress-testing discrete generative models and for developing evaluation metrics that detect real distributional mismatch rather than proxy performance. Our code is available at \url{https://github.com/shaojus/DiscreteDiffusionPlayground}.
\end{abstract}

\section{Introduction}

Discrete diffusion has emerged as a promising framework for generative modeling over categorical data, with applications ranging from natural language to biological sequences to molecular generation \cite{hoogeboom2021argmax,austin2021structured, avdeyev2023dirichlet, gruver2023protein,hoogeboom2022equivariant,dunn2024mixed}. Unlike continuous diffusion, which typically operates in Euclidean spaces \cite{sohl2015deep,ho2020denoising}, discrete diffusion is naturally aligned with domains in which each element belongs to a finite vocabulary. This makes it an appealing alternative to autoregressive generation, offering parallel decoding and flexible corruption processes while preserving the discrete structure of the underlying data \cite{lou2023discrete,sahoo2024simple}.

At the same time, evaluating discrete generative models remains difficult. The central goal of a generative model is to learn a distribution that matches the data distribution, yet this objective is rarely assessed directly in practice. Instead, prior work often relies on indirect metrics such as likelihood-based measures, perplexity, or embedding-based distances, as well as downstream task performance and qualitative inspection \cite{theis2015note}. These proxies can be useful, but they do not necessarily reveal whether the learned model faithfully captures the geometry, coverage, and mode structure of the target distribution. 

This challenge motivates a benchmark in which the target distribution is both explicit and interpretable. In this work, we introduce the \emph{Discrete Diffusion Playground}, a benchmark that starts from an analytically tractable distribution in \(\mathbb{R}^2\), quantizes each coordinate onto a finite grid, encodes the result as binary tokens, and trains discrete generative models purely on these sequences. Because the mapping between continuous points and discrete sequences is deterministic and bijective on the evaluation grid, generated samples can be decoded back into continuous space and compared directly against the known reference distribution. This construction preserves the discrete modeling problem while enabling visual and quantitative evaluation of distributional fidelity.

The low-dimensional structure of the Playground enables analyses that are difficult or impossible in realistic high-dimensional domains. We can compare decoded model samples with the target distribution using exact grid-based metrics, divergence measures, Wasserstein distance, sample-based tests, and kernel density visualizations \cite{villani2009optimal}. The benchmark also supports controlled variation in sequence layout, token order, and target distribution family, making it possible to study how discrete diffusion models respond to changes in inductive structure. Rather than replacing large-scale application benchmarks, we view the Playground as a complementary analysis tool: a transparent setting for comparing parameterizations, samplers, and evaluation metrics under conditions where model behavior can be diagnosed precisely.

Our main contributions are:
\begin{itemize}
    \item We introduce the \emph{Discrete Diffusion Playground}, a benchmark that links binary sequences to low-dimensional continuous distributions through an exact grid correspondence.
    \item We define controlled benchmark distributions, including Gaussian mixtures and piecewise-structured checkerboards, that expose smooth-density and sharp-support failure modes.
    \item We combine exact grid-based discrepancies with sample- and estimator-based metrics to directly assess distributional fidelity.
    \item We use this setting to compare autoregressive and discrete diffusion models, revealing failures that standard proxy metrics can obscure.
\end{itemize}


\section{Background}

Discrete diffusion models define generative processes over categorical data by learning to reverse a stochastic noising procedure. They are attractive for language and biological sequence generation, but their evaluation often relies on indirect metrics that do not fully capture whether the learned distribution matches the data \citep{theis2015note,borji2019pros}.

\subsection{Discrete Diffusion Models}

Diffusion models learn to reverse a noising process. Starting from clean data \(x\), the forward chain corrupts \(z_0=x\), and generation samples from a learned reverse chain:
\[
q(z_{1:T}\mid x)=\prod_{t=1}^T q(z_t\mid z_{t-1}),
\qquad
p_\theta(z_{0:T})=p(z_T)\prod_{t=1}^T p_\theta(z_{t-1}\mid z_t).
\]
In discrete diffusion, the states are categorical tokens and corruption is defined by vocabulary transitions. For a sequence \(x\in\mathcal{X}^L\), a forward step is often written using a categorical transition matrix \(Q_t\),
\[
q(z_t^{(i)}=b\mid z_{t-1}^{(i)}=a)=[Q_t]_{ab},
\qquad
q(z_t\mid z_{t-1})=\prod_{i=1}^L q(z_t^{(i)}\mid z_{t-1}^{(i)}).
\]

Early work extended diffusion to multinomial and structured categorical transitions \citep{hoogeboom2021argmax,austin2021structured}. A particularly important case is absorbing or masked diffusion, where tokens are independently replaced by a mask state \(m\) with probability \(\beta_t\):
\[
q(z_t^{(i)}=m\mid z_{t-1}^{(i)}\neq m)=\beta_t,
\qquad
q(z_t^{(i)}=z_{t-1}^{(i)}\mid z_{t-1}^{(i)}\neq m)=1-\beta_t.
\]
The reverse model reconstructs masked tokens from partially observed sequences, enabling non-autoregressive generation by revealing multiple positions in parallel. This flexibility comes with a harder learning problem: the model must solve many conditional prediction tasks over arbitrary masked subsets.

Recent methods differ mainly in how they parameterize and train this reverse process. SEDD estimates score ratios with a score-entropy objective \citep{lou2023discrete}; MDLM and MD4 simplify masked diffusion into cross-entropy-style denoising objectives \citep{sahoo2024simple,shi2024simplified}; and ReMDM modifies sampling through remasking dynamics \citep{wang2025remasking}. Our benchmark asks whether these variants recover the actual target distribution, not only whether they optimize proxy likelihoods or denoising losses.

\subsection{Evaluation of Generative Models}

Generative evaluation is difficult because the true data distribution is usually unknown, especially in high dimensions \citep{goodfellow2014generative}. Existing work therefore relies on proxies such as FID or Inception Score for images \citep{salimans2016improved,heusel2017gans}, perplexity for text \citep{brown1992class}, embedding distances, downstream performance, and qualitative samples. These measures are useful but indirect: they depend on pretrained feature spaces, sampling choices, or subjective inspection, and can miss mode dropping, spurious modes, off-support probability, or reduced diversity \citep{theis2015note,holtzman2019curious,hashimoto2019unifying}.
Recent work has similarly shown that discrete diffusion samplers can exhibit distributional error even with an oracle denoiser, and that improvements in standard likelihood- and sample-based metrics need not imply correct sampling \citep{tang2026your}. 
This motivates a benchmark with an explicit reference distribution and direct metrics for distributional mismatch.

\subsection{2D Discrete Quantization}
\label{app:2d_quantization}

Low-dimensional continuous distributions have previously been mapped to
discrete binary representations as controlled testbeds for generative
modeling. Dai et al.\ use 32-bit Gray-code representations of 2D
distributions to evaluate discrete energy-based models (EBMs), and
subsequent studies by Schröder et al.\ and Sun et al.\ adopt closely
related settings for discrete EBM and diffusion-model training
\citep{dai2020learning, schroder2023energy, sun2022score}.
More recently, Carter et al.\ use Gray-code discretizations of continuous
energy landscapes to evaluate discrete diffusion samplers
\citep{carter2026discrete}. Gray coding itself was originally introduced
by Gray \citep{gray1953pulse}. It represents consecutive integers using binary
strings that differ in exactly one bit, preserving adjacency between
neighboring quantization levels.

Our contribution is not the continuous-to-discrete encoding itself.
Rather, we build on this established evaluation protocol and provide an
accessible diagnostic framework for systematic comparison across model
families using analytically tractable target distributions. In contrast
to prior work, which primarily uses the protocol to validate a particular
EBM objective, diffusion model, or amortized sampler, we compare
autoregressive models with representative discrete diffusion models under
matched training conditions and evaluate them using both exact grid-based
and sample-based metrics.

We additionally include Gray-code encoding in the benchmark. The relative
ranking of the methods is highly robust to the encoding. We note that
there is a trend toward binary encoding being slightly more difficult for
some methods in the checkerboard setting, in particular SEDD.

\section{The Discrete Diffusion Playground}
\label{sec:playground}

The \emph{Discrete Diffusion Playground} evaluates discrete generative models through an exact correspondence with low-dimensional continuous data. Samples from an analytic distribution in \(\mathbb{R}^2\) are quantized, encoded as binary sequences, and used for discrete model training; generated sequences are decoded back to 2D for comparison with the reference distribution. This keeps the learning problem fully discrete while making failures such as missed modes, boundary blurring, or off-support mass directly visible.

\subsection{Overview of the construction}
\label{subsec:playground-overview}

\begin{figure}[!t]
    \centering
    \includegraphics[width=\linewidth, trim=0 80bp 0 80bp, clip]{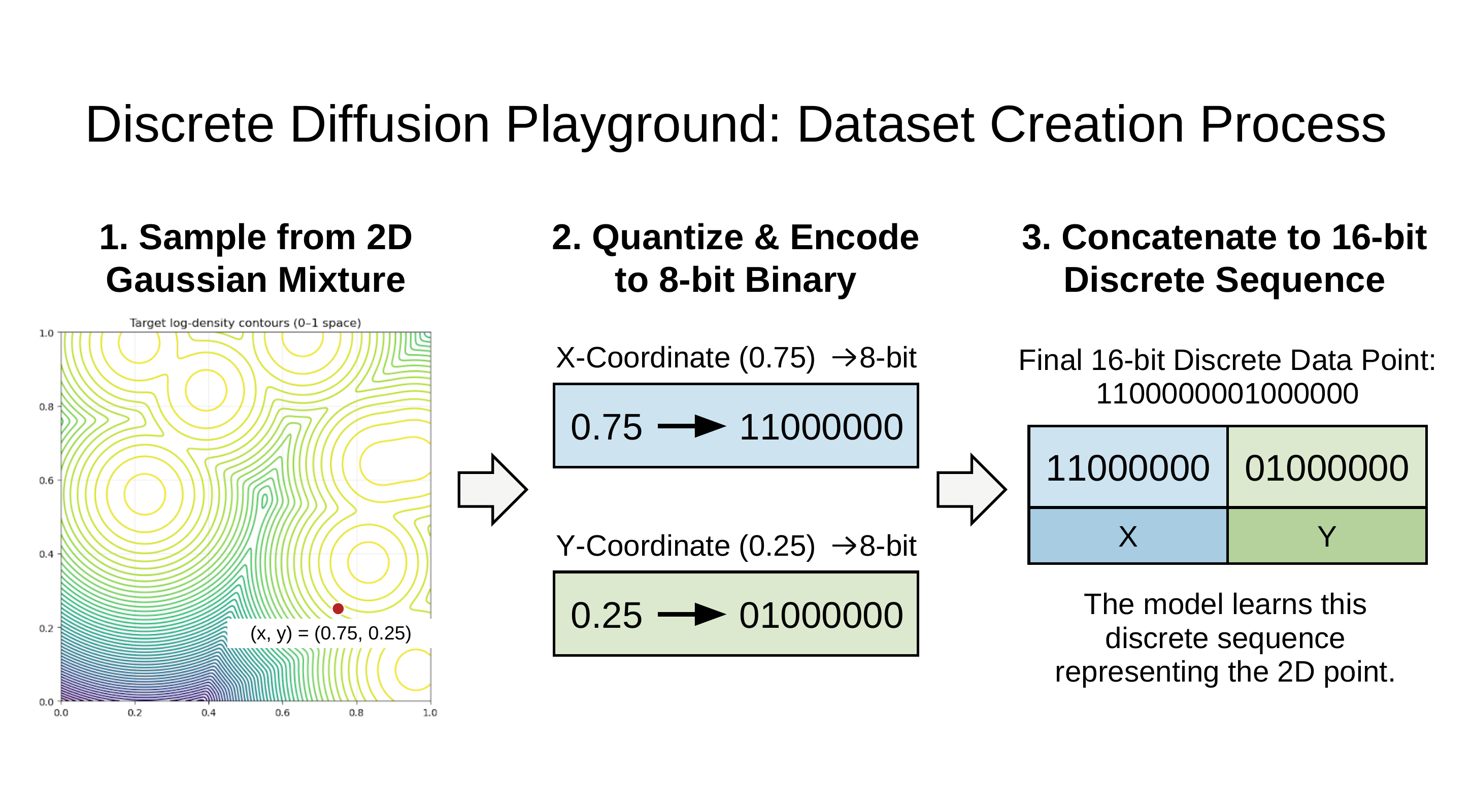}
    \caption{The Playground maps 2D samples to quantized binary sequences for discrete model training.}
    \label{fig:dataset-process}
\end{figure}

The benchmark consists of four stages:
\begin{enumerate}
    \item sample a point \((x,y)\) from a bounded continuous distribution;
    \item quantize each coordinate onto a finite grid;
    \item encode the quantized coordinates as binary strings;
    \item concatenate the coordinate-wise encodings into a sequence for model training.
\end{enumerate}
Figure~\ref{fig:dataset-process} shows an 8-bit-per-coordinate example. A sampled 2D point is quantized into integer grid coordinates, each coordinate is written as an 8-bit binary string, and the two strings are combined into a 16-token sequence. The model only sees these binary sequences during training, but each valid sequence decodes to a unique point on the underlying 2D grid.

\paragraph{Why binary tokens.}
Binary tokens keep the state space simple while assigning each token a clear geometric role: more significant bits determine coarse location, and less significant bits refine it. This makes token position interpretable, enabling controlled studies of order sensitivity, masking behavior, and sampling trajectories \citep{hoogeboom2021argmax,austin2021structured}.

\paragraph{Sequence layout.}
The default layout concatenates all \(x\) bits followed by all \(y\) bits. We also consider interleaved orderings,
\[
(x_1, y_1, x_2, y_2, \dots, x_b, y_b),
\]
and reversed variants. These layouts preserve the same data distribution but change the inductive structure presented to the model.

\paragraph{Autoregressive order controls.}
Because autoregressive likelihood depends on sequence factorization, we compare a coarse-to-fine ordering with a reversed/interleaved ordering that disrupts the natural bit hierarchy. This tests whether AR performance is simply an artifact of favorable token order.

\subsection{Supported distribution families}
\label{subsec:playground-distributions}

The Playground supports continuous distributions that can be sampled efficiently on a bounded domain and evaluated through either a density or a high-quality reference mass function.

\paragraph{Mixtures of Gaussians.}
Our primary smooth benchmark is a 2D Gaussian mixture,
\begin{equation}
p(\mathbf{z}) = \sum_{k=1}^{K} \pi_k \, \mathcal{N}(\mathbf{z}; \mu_k, \Sigma_k), \qquad \mathbf{z}=(x,y),
\end{equation}
with user-specified weights, means, and covariances. Gaussian mixtures provide controllable multimodality, tunable overlap, closed-form PDFs, and interpretable visualizations.

\paragraph{Checkerboards and piecewise-structured densities.}
Checkerboards emphasize sharp boundaries, disconnected support, and local combinatorial structure. They test whether models can avoid invalid regions rather than merely approximate smooth density contours.

\paragraph{Other admissible distributions.}
Any bounded continuous distribution can be used if samples are efficient to draw and an evaluation reference is available.

\subsection{Evaluation Metrics}
\label{subsec:playground-metrics}

Because the target distribution is explicit and low-dimensional, the Playground supports direct evaluation against a known reference. We compare decoded model samples to the target both on an evaluation grid and in continuous space.

\subsubsection{Exact grid-based metrics}

Let \(P_t\) be the target mass function on an evaluation grid and \(P_m\) the model-induced mass function from decoded samples. Once both are represented on the same grid, discrepancies can be computed directly. The forward Kullback--Leibler divergence \citep{kullback1951information} is
\begin{equation}
\mathrm{KL}(P_t \,\|\, P_m)
=
\sum_i P_t(i)\log \frac{P_t(i)}{P_m(i)}.
\end{equation}
Forward KL penalizes missing target mass and is therefore sensitive to mode dropping or poor coverage. Reverse KL penalizes model mass in low-density or unsupported regions, making it useful for detecting spurious samples and support leakage. We also compute MSE between cell masses,
\begin{equation}
\mathrm{MSE}(P_t,P_m) = \frac{1}{N}\sum_i \bigl(P_t(i)-P_m(i)\bigr)^2,
\end{equation}
where \(N\) is the number of grid cells, along with maximum cell-wise error and exact 2D Wasserstein distance on the shared discrete grid. Because MSE averages over the full grid, it can understate localized but distributionally important errors, such as mass in invalid cells.

\subsubsection{Sample- and estimator-based metrics}

We additionally report sample-based metrics on decoded points, including the energy distance statistic \citep{szekely2013energy}. Kernel density estimates provide qualitative diagnostics for spurious bridges, boundary artifacts, and oversmoothing \citep{silverman2018density}, but are treated as supportive because they depend on bandwidth selection.

\subsection{Models evaluated}
\label{subsec:models-evaluated}

We evaluate one autoregressive baseline and four representative discrete diffusion models. The autoregressive transformer (AR) uses a fixed left-to-right factorization \citep{vaswani2017attention}. SEDD estimates score ratios through a score-entropy objective \citep{lou2023discrete}. MDLM uses a simplified masked diffusion language-modeling formulation \citep{sahoo2024simple}. MD4 trains a masked discrete diffusion model with a simplified cross-entropy objective \citep{shi2024simplified}. ReMDM modifies masked diffusion sampling through remasking dynamics \citep{wang2025remasking}. All models use Transformer-based sequence architectures, with capacity scaled by varying the model dimension and number of layers. We evaluate 11 configurations for the GMM experiments and six for the checkerboard experiments; full details are provided in Appendix~\ref{app:hyperparameters}.

\subsection{Scope and limitations}
\label{subsec:playground-limitations}

The Playground is a diagnostic benchmark, not a replacement for high-dimensional application benchmarks. Its binary sequences are synthetic and more structured than natural language, protein sequences, or molecular graphs. This simplicity is intentional: it isolates representation, ordering, sampling, and metric behavior in a setting where the ground truth is known.

\section{Results}
\label{sec:results}

\vspace{-3pt}

We primarily compare autoregressive and masked diffusion models on binary-encoded 2D distributions. Generated sequences are decoded into normalized coordinates and evaluated with sample visualizations and distributional metrics against the known target. We emphasize distributional fidelity rather than likelihood alone: reasonable token probabilities can still produce samples that miss modes, blur boundaries, or place mass in invalid regions. We additionally repeat the comparison using Gray-code representations and
observe the same overall model ranking and qualitative trends, suggesting
that the main conclusions are not specific to standard binary encoding
(Appendix~\ref{app:gray-code-encoding}).

\vspace{-1pt}

\subsection{Autoregressive models achieve stronger distributional fidelity on the Gaussian mixture benchmark}
\label{subsec:gmm-scaling}

\vspace{-1pt}

\begin{figure}[tb]
    \centering
    \includegraphics[width=0.95\textwidth]
    {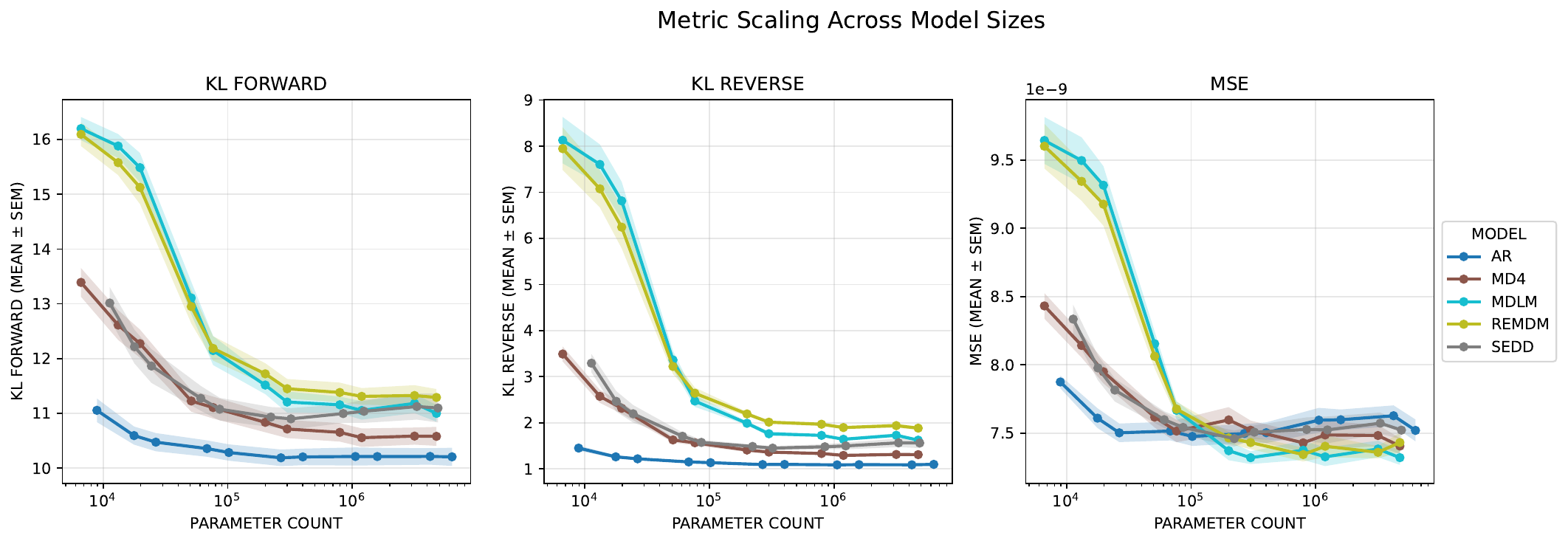}
    \caption{Scaling on the Gaussian mixture benchmark for forward KL, reverse KL, and MSE. AR Reversed is analyzed separately. Lines show means over 10 independent seeds (0–9), and shaded regions indicate $\pm$ one standard error of the mean (SEM).}
    \label{fig:gmm-kl-mse-scaling}
\end{figure}

We first evaluate the smooth multimodal Gaussian mixture benchmark, where the target distribution is exactly evaluable on a grid. Figure~\ref{fig:gmm-kl-mse-scaling} shows forward KL, reverse KL, and MSE across model families, excluding AR Reversed so the main comparison reflects distinct model classes rather than an additional autoregressive ordering control. All methods improve from small to intermediate parameter counts, showing that the task is not saturated at very low capacity. However, AR reaches the most stable low-KL regime, while masked diffusion models remain higher and more variable.

The KL metrics show the clearest separation. Forward KL reflects target coverage, while reverse KL reflects off-mode or low-density model mass. AR is strongest on both, suggesting that its fixed factorization fits the coarse-to-fine binary coordinate representation: early bits determine broad spatial region, and later bits refine local position. MSE is less separated: several masked diffusion models become competitive in MSE even when their KL values remain worse. Thus, similar MSE does not imply similar generative fidelity: KL more strongly penalizes missing target mass and probability assigned to low-density regions.

\begin{figure*}[tb]
    \centering

    \includegraphics[width=0.9\textwidth]{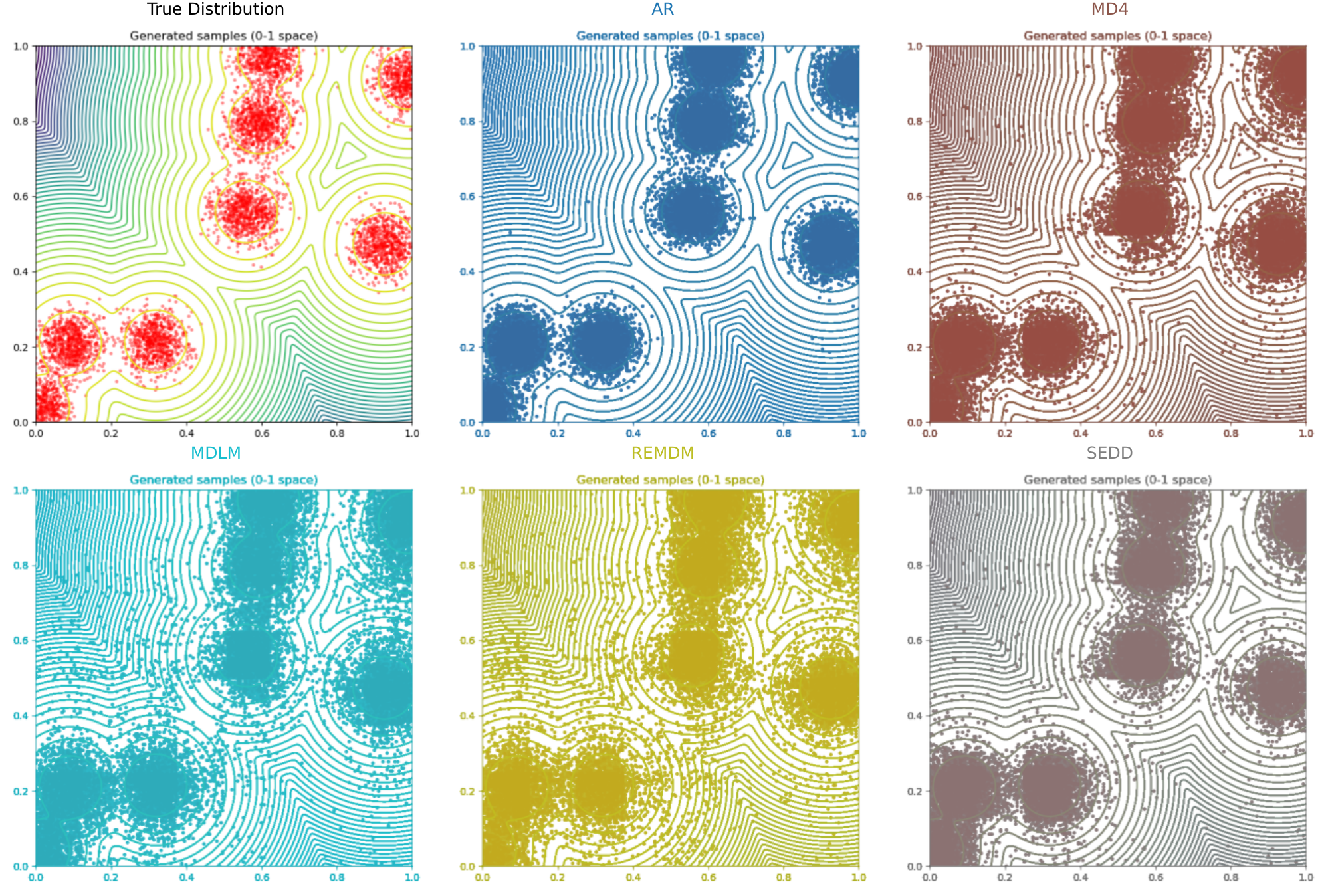}

    \vspace{1ex}

    \includegraphics[width=0.9\textwidth]{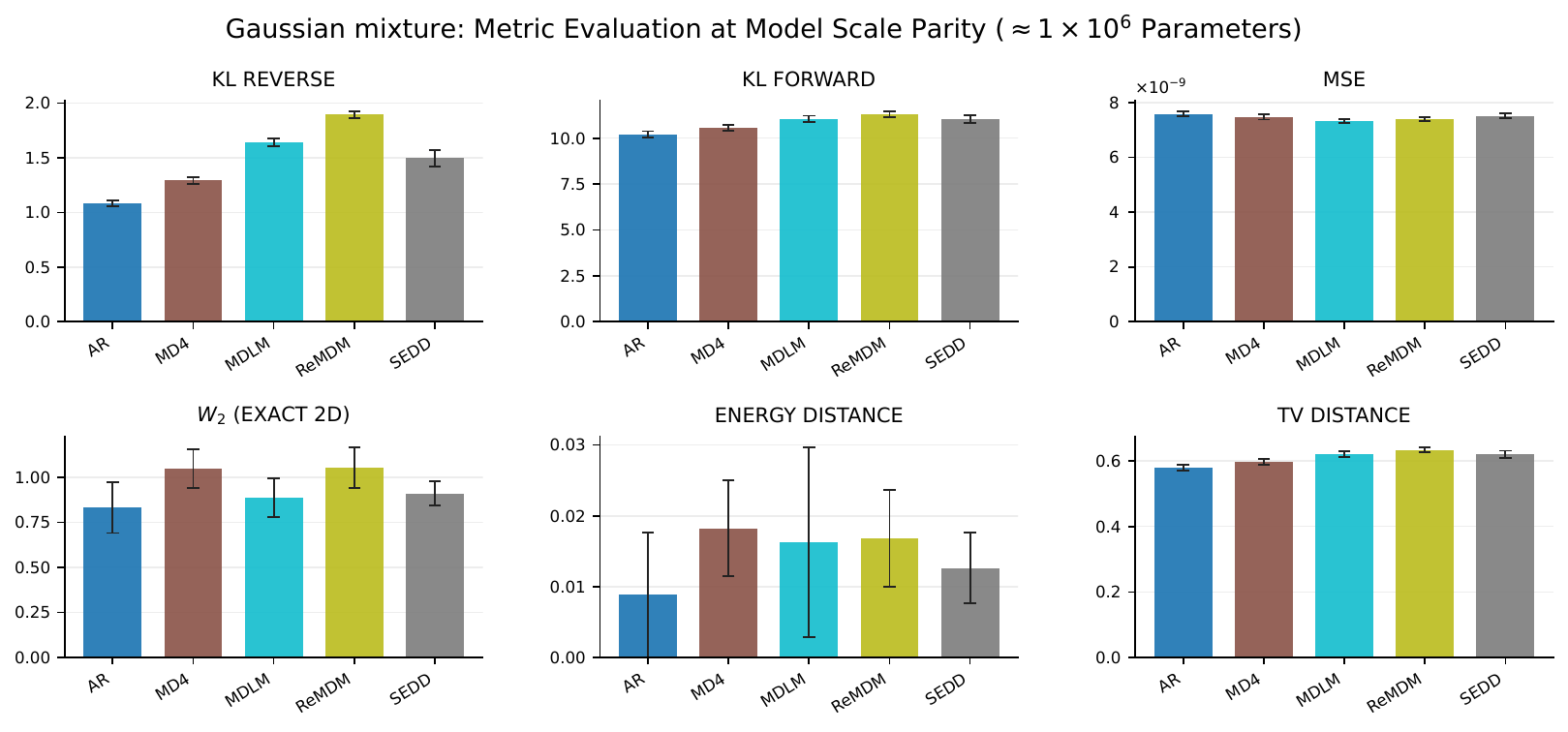}

    \caption{Scale-parity comparison on the Gaussian mixture benchmark. Top: decoded samples overlaid on contours of the target Gaussian mixture density at a comparable model budget. Bottom: metric comparison at approximately \(10^6\) parameters. Bars show means over 10 seeds (0–9), with error bars indicating $\pm$ 1 SEM.
    }
    \label{fig:gmm-scale-parity}
\end{figure*}

Figure~\ref{fig:gmm-scale-parity} summarizes the matched-scale comparison. At approximately \(10^6\) parameters, all models recover the coarse GMM structure, but AR produces sharper samples with less visible mass between modes. Masked diffusion models capture dominant modes but are more diffuse near component boundaries. These subtle visual differences are amplified by reverse KL, TV, Wasserstein distance, and energy distance, so we treat MSE as a useful density-error diagnostic but not the primary measure of generative quality.

This illustrates why the Playground is useful even on smooth targets. Models that recover the approximate sample cloud can still differ in mass between modes, low-probability coverage, or local smoothing. Forward KL, reverse KL, and transport-style metrics expose complementary aspects of these errors.

\vspace{-1pt}

\subsection{Reversing autoregressive bit order produces only a small early training delay}
\label{subsec:ar-reversed-ordering}

\paragraph{Autoregressive order controls.}
For autoregressive models, sequence layout is part of the factorization and could therefore create an artificial advantage if the tokens are presented in their most natural order. We therefore evaluate both a best-case and a worst-case autoregressive ordering. The best-case ordering, AR, presents the binary coordinates in a coarse-to-fine layout, so that early tokens determine broad spatial location and later tokens refine position. The worst-case ordering, AR Reversed, reverses and interleaves the bits, disrupting this coarse-to-fine structure and forcing the model to learn dependencies under a less geometrically aligned factorization. Comparing these settings allows us to distinguish genuine autoregressive performance from an artifact of a favorable token order.
We analyze AR Reversed separately because it tests whether AR's advantage is merely caused by the favorable coarse-to-fine bit layout.

\begin{figure}[tb]
    \centering
    \includegraphics[width=\linewidth]{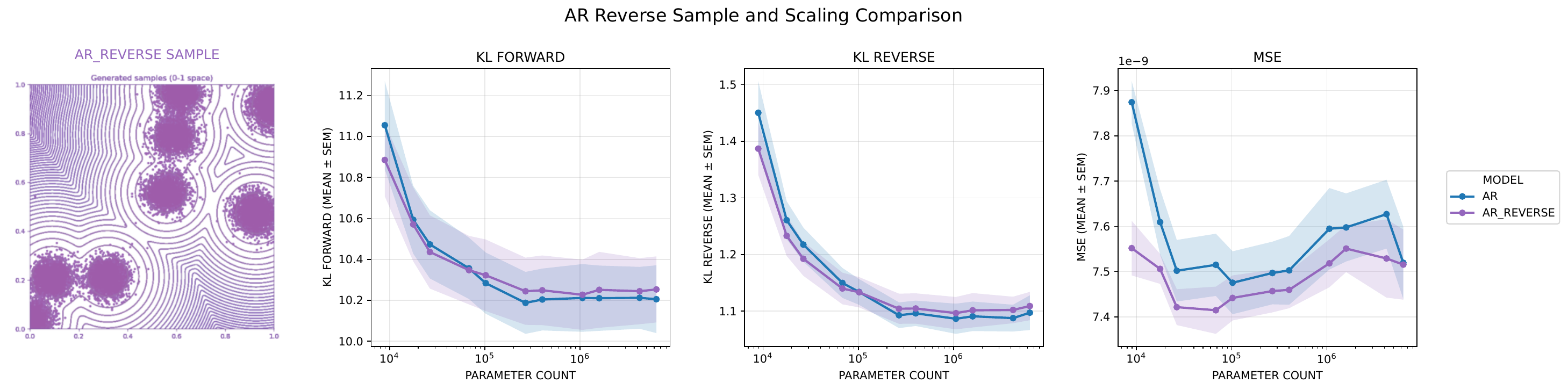}
    \caption{AR Reversed ordering control on the Gaussian mixture benchmark. Left: representative samples. Right: scaling comparison with standard AR. Lines show means over 10 seeds (0–9), and shaded regions indicate $\pm$ 1 SEM.}
    \label{fig:ar-reversed-discussion}
\end{figure}

Figure~\ref{fig:ar-reversed-discussion} shows that the expected ordering penalty does not persist as a large final performance gap. AR Reversed produces samples close to the target GMM, and its scaling curves closely track standard AR. This is especially clear for reverse KL: although reversing the order could have increased off-mode probability, AR Reversed remains close to AR across most model sizes. 

\begin{figure}[tb]
    \centering
    \includegraphics[width=\linewidth]{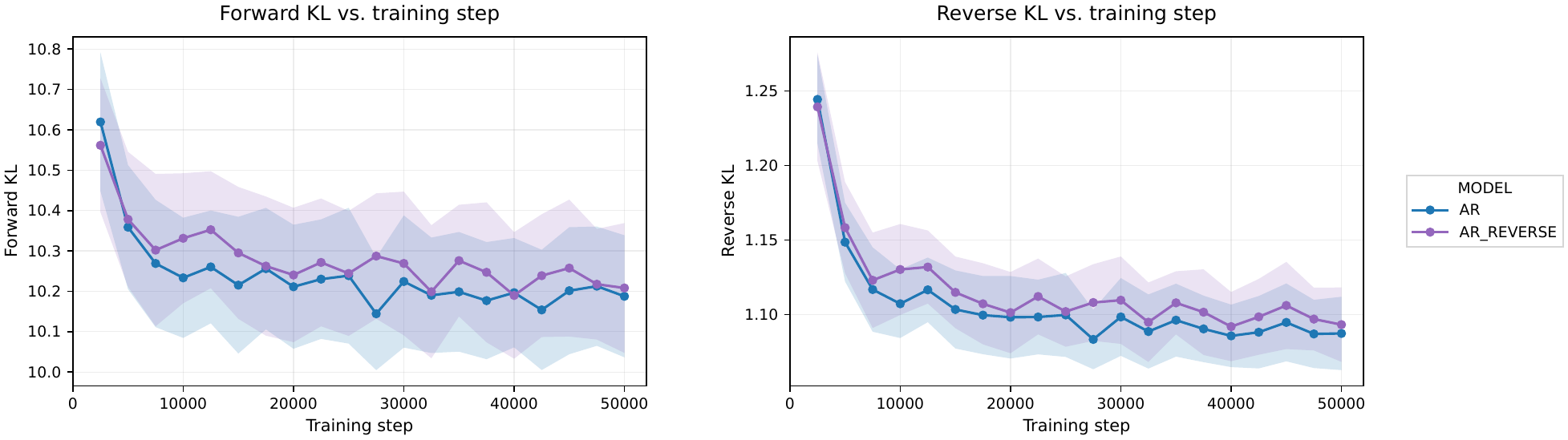}
    \caption{Effect of sequence ordering on autoregressive training. AR and AR Reversed converge similarly. Lines show means over 10 seeds (0–9), and shaded regions indicate $\pm$ 1 SEM.}
    \label{fig:ordering-training}
\end{figure}

\vspace{-5pt}

Figure~\ref{fig:ordering-training} shows a short-lived ordering effect: the in-order model achieves slightly lower forward and reverse KL early in training, consistent with a small sample-efficiency advantage from the coarse-to-fine layout. However, the gap narrows with sufficient optimization. Thus, AR's advantage over masked diffusion is not simply caused by a favorable bit ordering; even under a disrupted order, the autoregressive transformer can learn dependencies that map back onto the same two-dimensional geometry.

\subsection{Masked models learn structured reveal orders under adaptive sampling}
\label{subsec:learned-reveal-order}

Beyond final distributional metrics, the Playground also lets us inspect how a masked generative model assembles a binary-encoded sample. We use adaptive top-$k$ reveal sampling as a diagnostic: at each step, the model predicts all currently masked positions, and the most confident positions are revealed first. This produces an empirical reveal-time distribution for each bit position, allowing us to ask whether masked models recover meaningful generation orders from the learned conditionals rather than from an explicitly imposed autoregressive factorization.

Supplementary Figure~\ref{fig:learned-reveal-order-supp} shows that the resulting reveal schedules align with the structure of the binary representation across standard, interleaved, and reversed layouts. In the standard layout, high-order coordinate bits tend to be revealed earlier than low-order refinement bits, consistent with a coarse-to-fine generation strategy. In the interleaved layout, paired $x$ and $y$ bits are revealed at similar stages, suggesting that the schedule adapts to the coupled spatial role of the two coordinates. In the reversed layout, the pattern changes with the token order, indicating that adaptive sampling can recover representation-dependent structure rather than simply following a fixed left-to-right bias.

This diagnostic highlights a use of the Playground beyond aggregate sample quality evaluation. Because each token has a known geometric role, reveal-time analysis makes it possible to visualize whether a masked model's sampling procedure aligns with the latent structure of the data. This provides a complementary view of model behavior alongside KL, Wasserstein distance, and MSE.

\subsection{Rotated checkerboards expose sharper support-matching failures}
\label{subsec:checkerboard-four}

Because smooth GMM samples can look plausible despite mass between modes, we next evaluate a rotated and shifted \(4\times4\) checkerboard with sharp occupied and empty regions. This directly tests support matching.

\begin{figure}[tb]
    \centering
    \includegraphics[width=0.95\textwidth]{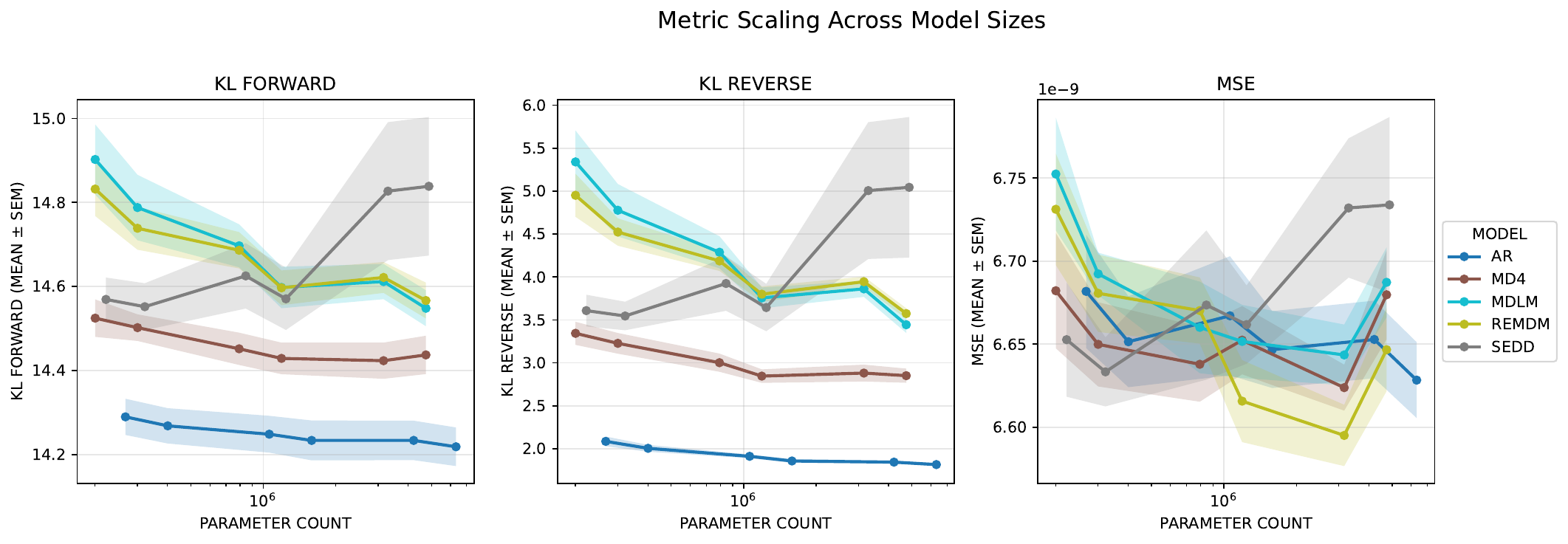}
    \caption{Metric scaling on the rotated and shifted 4×4 checkerboard benchmark. Lines show means over 10 seeds (0–9), and shaded regions indicate $\pm$ 1 SEM.}
    \label{fig:checkerboard-scaling}
\end{figure}

Figure~\ref{fig:checkerboard-scaling} shows that AR again has the best KL behavior, especially reverse KL, indicating less probability in invalid regions. Masked diffusion models scale less consistently; MD4, MDLM, and ReMDM remain close to one another, while SEDD degrades most noticeably at larger sizes. This suggests that increasing parameter count alone does not resolve the denoising and sampling failures of discrete diffusion models on sharply structured supports.

\begin{figure*}[tb]
    \centering

    \includegraphics[width=0.9\textwidth]{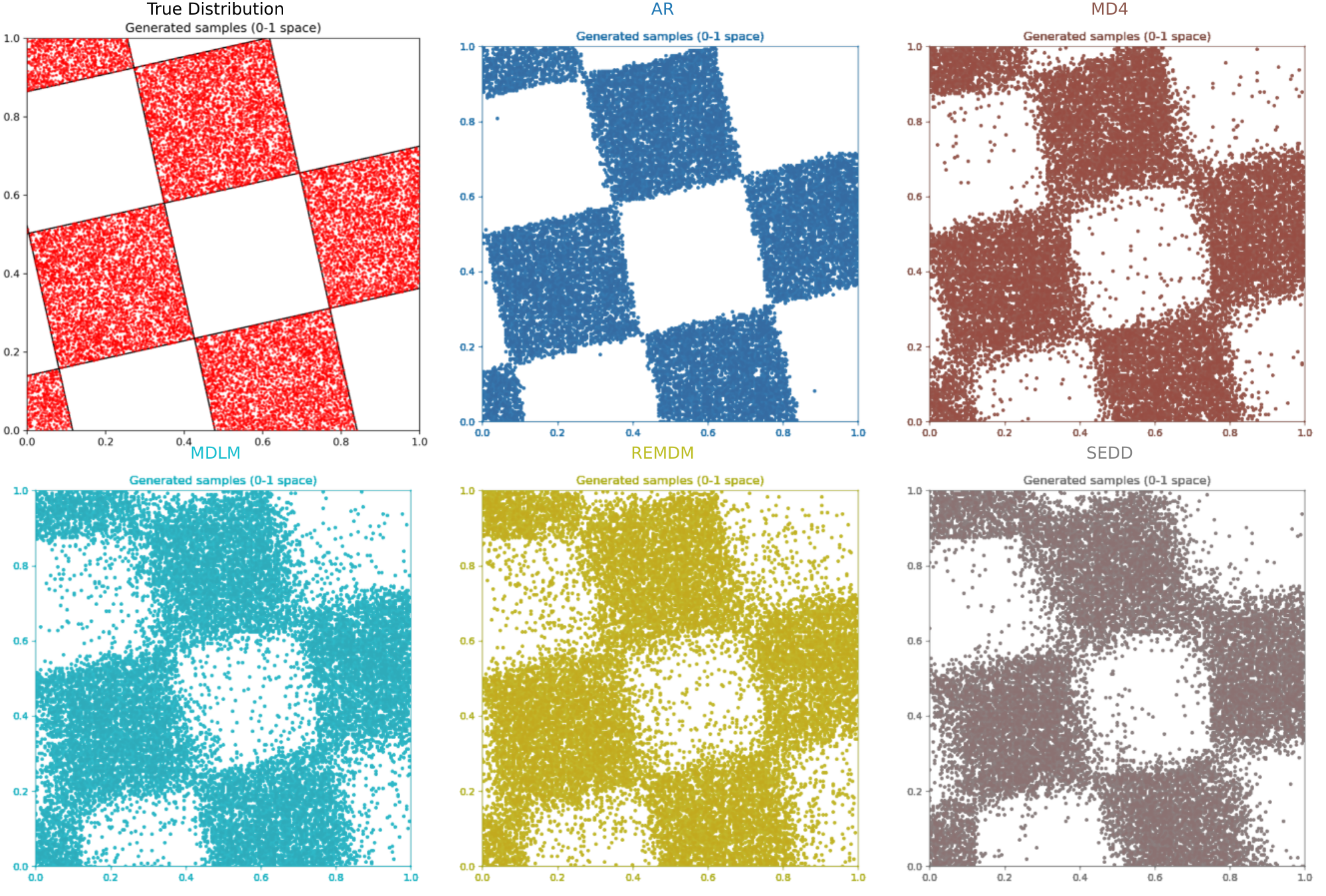}

    \vspace{1ex}

    \includegraphics[width=0.9\textwidth]{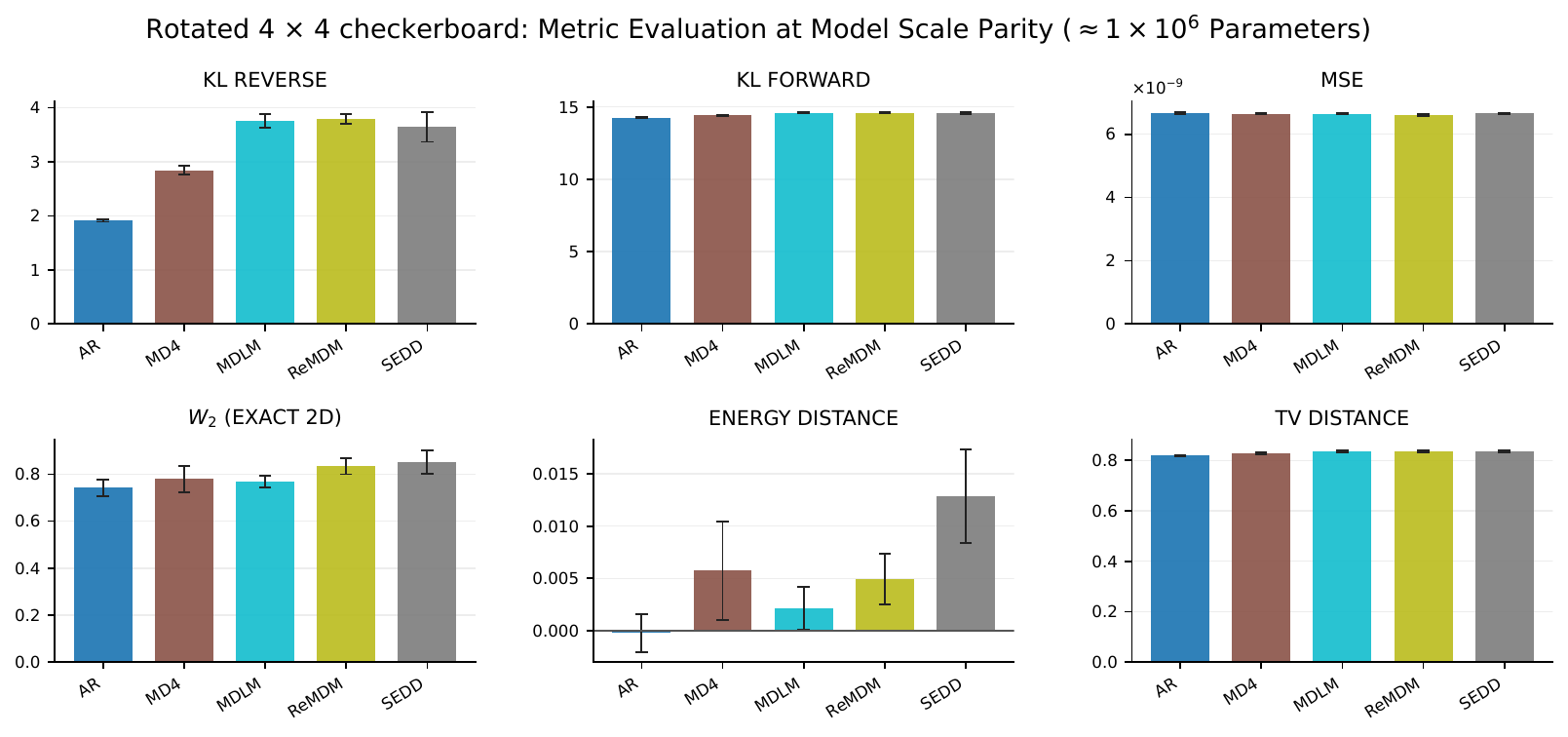}

    \caption{Scale-parity comparison on the rotated and shifted 4×4 checkerboard benchmark. Top: representative generated samples at approximately matched model scale. Bottom: matched-scale metric comparison. Bars show means over 10 seeds (0–9), with error bars indicating $\pm$ 1 SEM.
    }
    \label{fig:checkerboard-combined}
\end{figure*}

Figure~\ref{fig:checkerboard-combined} explains the metric gap. AR most closely recovers the rotated checkerboard support, producing samples concentrated in the occupied cells with relatively sharp separation from empty regions. MD4, MDLM, and ReMDM capture coarse structure but blur across boundaries and place visible mass in invalid regions, while SEDD shows the weakest boundary fidelity. Reverse KL and TV respond strongly to these errors because both penalize probability assigned outside the target support. MSE remains similar across models because many cells contribute only small absolute errors, illustrating why average grid-wise squared error can be misleading on sparse structured distributions.

The rotated checkerboard cannot be solved by matching independent coordinate marginals: occupancy depends on joint coordinate structure after rotation and shift. Thus, locally plausible token predictions can still decode to invalid cells, making support-based diagnostics essential.

\vspace{-1pt}

\section{Discussion}
\label{sec:discussion-limitations}

Across the Gaussian mixture and checkerboard benchmarks, autoregressive models are a strong baseline for binary-encoded geometric data. Their fixed factorization is well matched to coordinate encodings, where early bits determine broad location and later bits refine it. This conclusion comes from decoded distributional behavior rather than token-level likelihood alone.

Masked diffusion models improve with scale but remain less stable across metrics. Their flexibility supports adaptive inference, but the checkerboards show that plausible smooth-target samples can still coexist with probability leakage across sharp boundaries. Better denoising loss or perplexity therefore does not guarantee better decoded geometry.

This result should not be read as a general rejection of masked diffusion. The Playground removes advantages masked models may have in realistic settings, such as infilling. Instead, it asks a narrower question: \emph{when the target is known and visualizable, which models reproduce its geometry most faithfully?}

The ordering experiments show that sequence layout affects optimization more than final learnability, so representation and ordering should be treated as part of benchmark design rather than incidental preprocessing choices.

The Playground is intentionally simple and does not predict performance in high-dimensional language, protein, or molecular domains by itself. Its value is diagnostic: known token semantics and explicit targets isolate failures in representation, ordering, sampling, and evaluation metrics.

A practical use of this benchmark is therefore as an intermediate debugging layer before biological sequence or molecular design tasks where failures are hard to localize. If a method cannot preserve rotated checkerboard support or assigns probability between GMM modes, those failures can guide changes to masking schedules, sampling rules, model parameterization, or evaluation metrics. Success on the Playground does not prove real-world performance, but it provides a more interpretable check than proxy likelihood or qualitative samples alone.

\section{Conclusion}
\label{sec:conclusion}

We introduced the Discrete Diffusion Playground, a 2D benchmark that maps analytic continuous distributions to binary sequences while enabling direct evaluation against known targets. Across exact grid divergences, transport metrics, sample tests, and visual diagnostics, autoregressive models learned these distributions more reliably than masked diffusion variants, especially on rotated checkerboards with sharp support constraints. The Playground provides a compact testbed for debugging discrete generative models and developing metrics that detect true distributional mismatch rather than proxy performance. An interesting open question is whether similarly simple settings can be constructed in which distributional fidelity is directly evaluable and discrete diffusion exhibits a clear advantage over autoregressive modeling.

\begin{ack}
This work was supported in part by the National Institutes of Health (NIH) under grant R01HG013724 and R35GM140753 and the National Science Foundation (NSF) under grant DBI-2238125.
This work used Bridges-2 at the Pittsburgh Supercomputing Center through allocation cis260238p from the Advanced Cyberinfrastructure Coordination Ecosystem: Services \& Support (ACCESS) program, which is supported by National Science Foundation grants 2138259, 2138286, 2138307, 2137603, and 2138296.
\end{ack}

{\small
\bibliographystyle{plainnat}
\bibliography{playground}
}

\FloatBarrier

\clearpage
\appendix

\setcounter{figure}{0}
\renewcommand{\thefigure}{S\arabic{figure}}

\section{Supplementary Results}
\label{sec:supplementary-results}

\subsection{Learned reveal-order diagnostics}
\label{app:learned-reveal-order}

\begin{figure}[!htbp]
    \centering
    \includegraphics[width=0.49\linewidth]{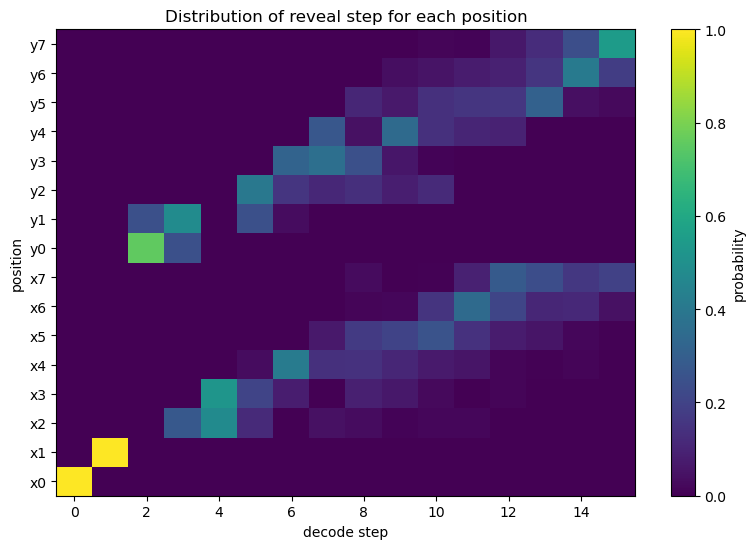}
    \includegraphics[width=0.49\linewidth]{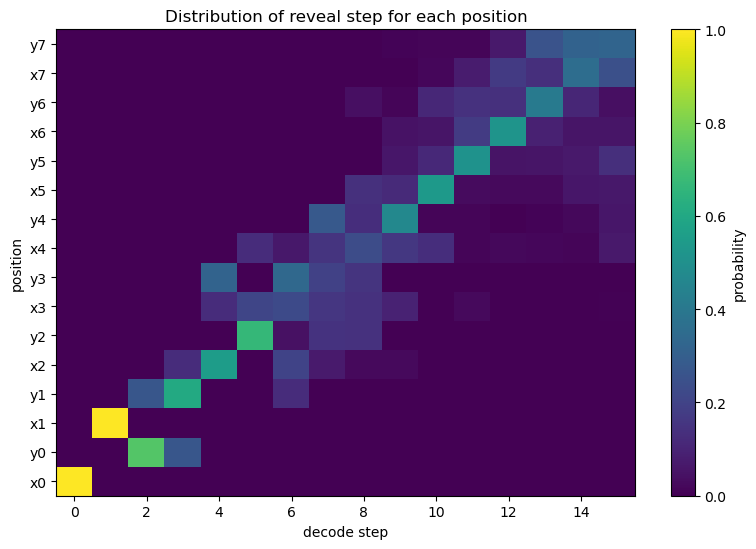}

    \includegraphics[width=0.62\linewidth]{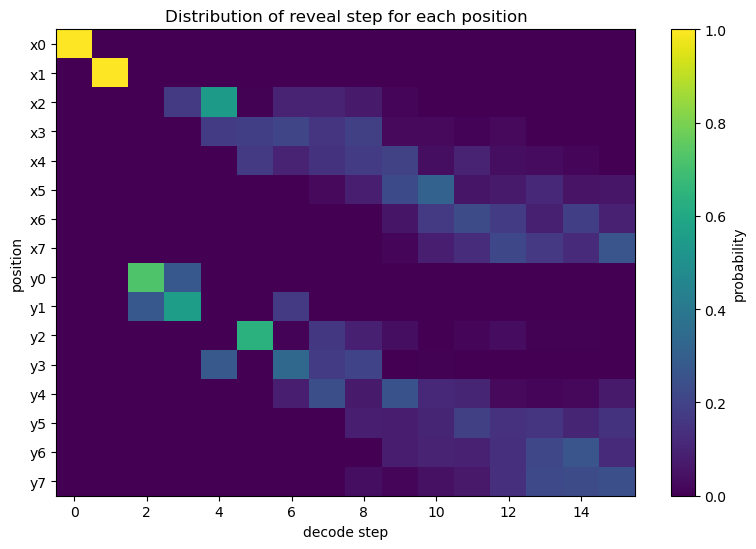}
    \caption{
    Learned reveal order under top-$k$ adaptive masked sampling. Each heatmap shows the probability that a bit position is revealed at each decoding step. Although the model is trained with random masking, adaptive sampling recovers structured reveal schedules for standard, interleaved, and reversed sequence layouts. Top left: standard layout; top right: interleaved layout; bottom: reversed layout.
    }
    \label{fig:learned-reveal-order-supp}
\end{figure}

Figure~\ref{fig:learned-reveal-order-supp} provides the full reveal-order diagnostics discussed in Section~\ref{subsec:learned-reveal-order}.

\subsection{The \(8\times8\) checkerboard amplifies masked diffusion failures}
\label{subsec:checkerboard-eight}

We further increase the difficulty by evaluating a rotated and shifted \(8\times8\) checkerboard. This distribution contains more occupied regions, smaller cells, and more frequent alternations between high-density and zero-density regions. A model must recover the global rotation and shift while also preserving many fine-scale local boundaries.

\begin{figure}[!htbp]
    \centering
    \includegraphics[width=0.9\linewidth]{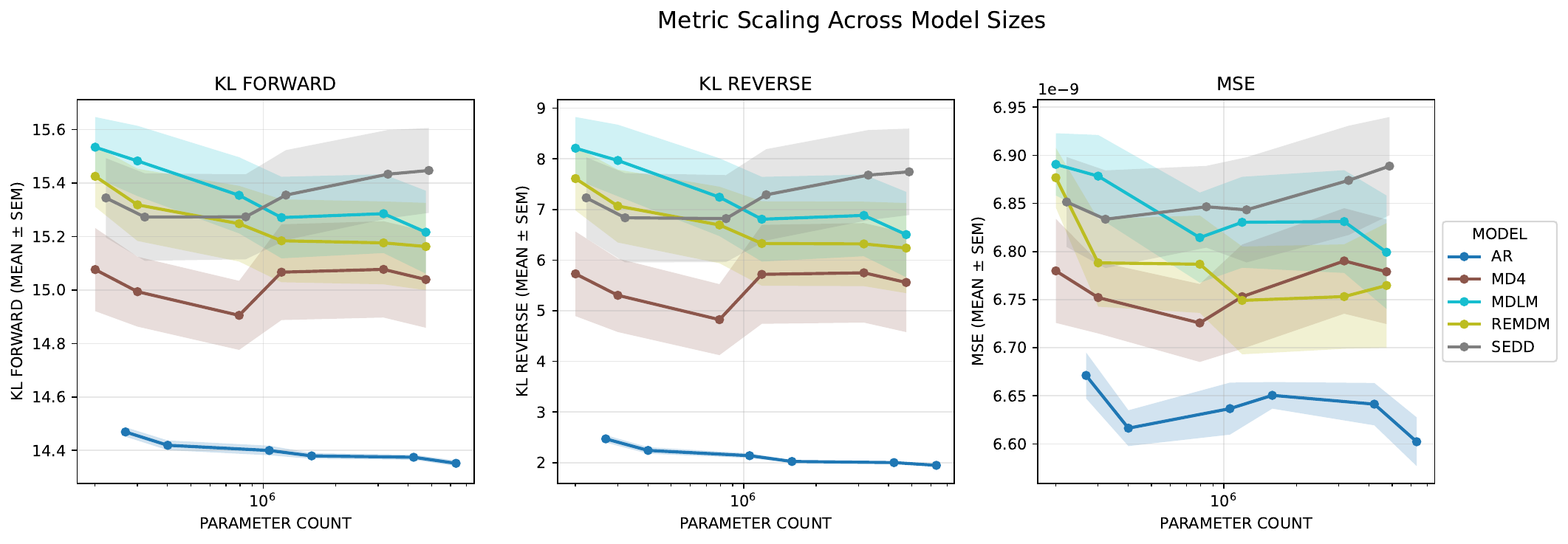}
    \caption{
    Metric scaling on the rotated and shifted \(8\times8\) checkerboard benchmark. Lines show means over 10 seeds (0–9), and shaded regions indicate $\pm$ 1 SEM.
    AR improves with model scale and achieves the strongest KL forward and KL reverse performance.
    Masked diffusion models show weaker and less consistent scaling, while SEDD remains near the worst KL values across model sizes.
    MSE provides a weaker separation between models than the KL metrics. 
    }
    \label{fig:checkerboard-eight-scaling}
\end{figure}

Figure~\ref{fig:checkerboard-eight-scaling} shows that AR remains the strongest model across the KL metrics. Its KL values improve with scale, indicating better mode coverage and reduced off-support mass. The masked diffusion models show weaker scaling. MD4, MDLM, and ReMDM remain substantially worse than AR, while SEDD performs worst overall and does not recover with increased model size. This indicates that the failure is not simply due to insufficient capacity.

\begin{figure}[!htbp]
    \centering
    \includegraphics[width=0.9\linewidth]{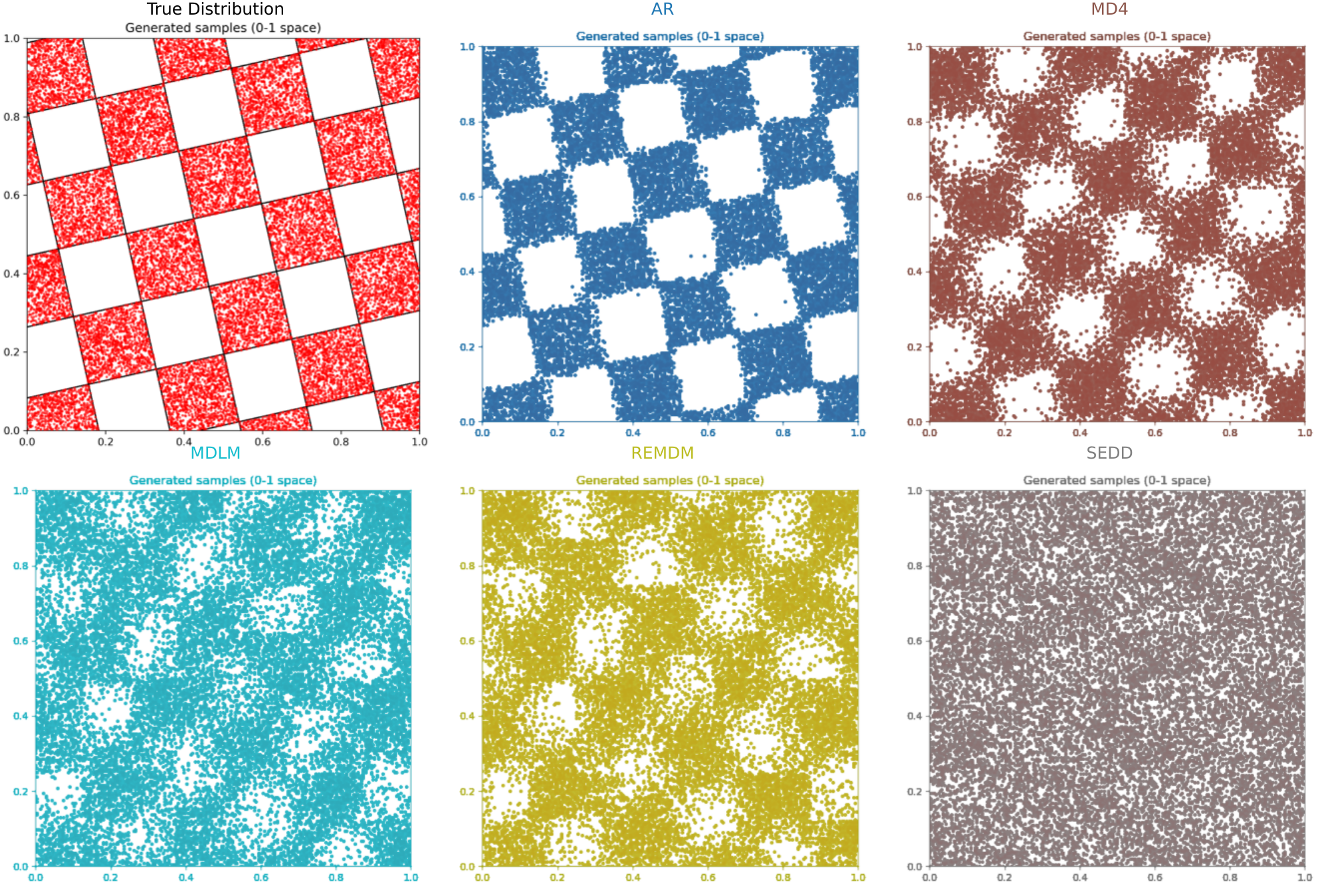}
    \caption{
    Generated samples on the rotated and shifted \(8\times8\) checkerboard benchmark.
    AR best preserves the alternating checkerboard structure.
    MD4, MDLM, and ReMDM show increasing levels of diffusion across cell boundaries.
    SEDD largely fails to recover the target support and produces samples that are close to uniformly spread over the domain.
    }
    \label{fig:checkerboard-eight-samples}
\end{figure}

The sample visualizations in Figure~\ref{fig:checkerboard-eight-samples} make the failure mode more apparent. AR preserves the repeated rotated checkerboard pattern and assigns most samples to the correct occupied cells. MD4 retains some coarse structure but blurs across boundaries. MDLM and ReMDM spread samples more broadly across the domain. SEDD fails most severely, producing samples that appear close to globally diffuse with little visible recovery of the target support. Since the true \(8\times8\) checkerboard contains many empty regions, this kind of broad probability leakage is heavily penalized by reverse KL and TV distance.

\begin{figure}[!htbp]
    \centering
    \includegraphics[width=0.92\linewidth]{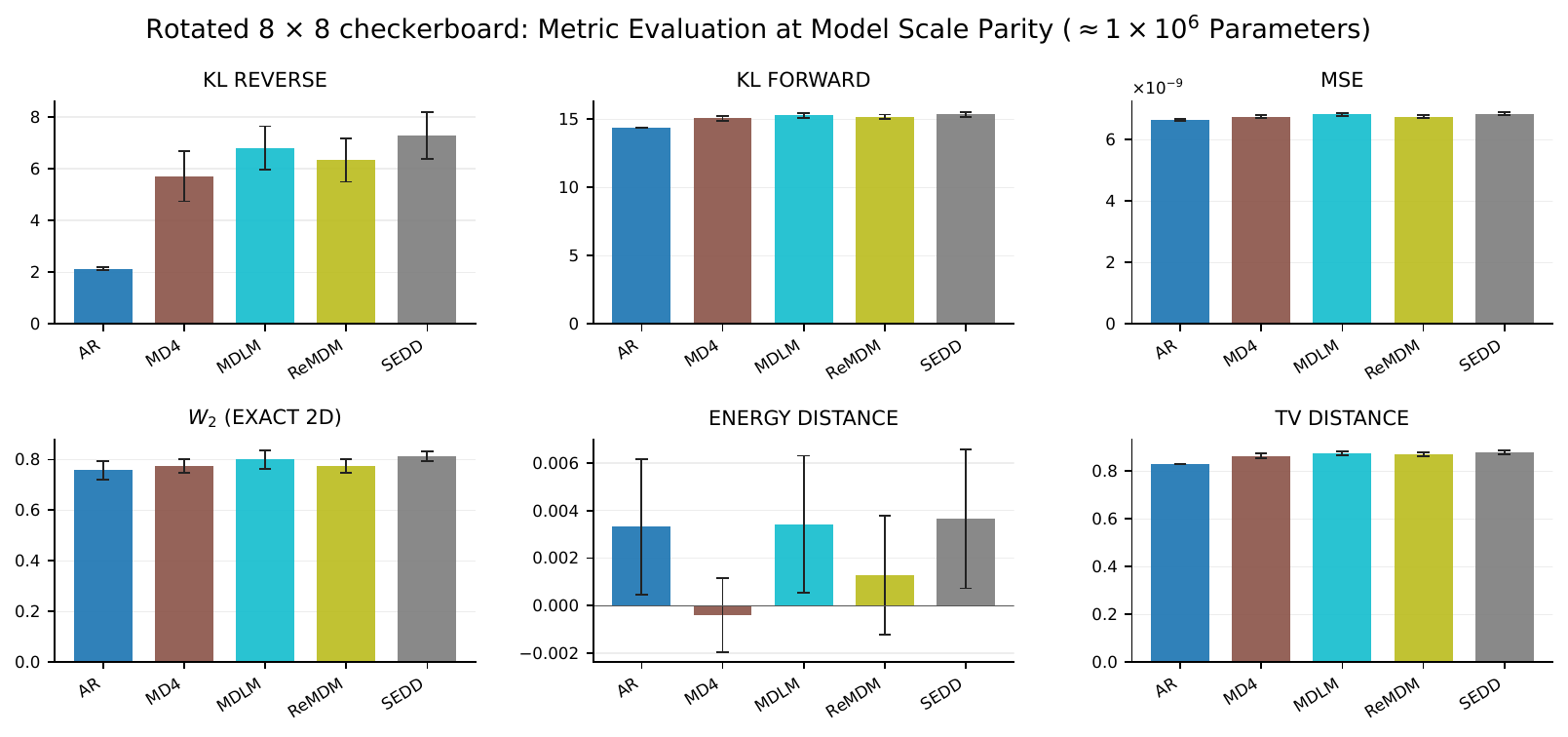}
    \caption{
    Metric comparison at approximately matched model scale on the rotated and shifted \(8\times8\) checkerboard benchmark. Bars show means over 10 seeds (0–9), with error bars indicating $\pm$ 1 SEM.
    AR achieves the strongest performance on KL reverse, KL forward, Wasserstein distance, energy distance, and TV distance.
    MSE remains similar across models despite large qualitative differences in generated samples, again showing that average grid-wise squared error is not sufficient for evaluating sparse structured distributions.
    }
    \label{fig:checkerboard-eight-budget}
\end{figure}

Figure~\ref{fig:checkerboard-eight-budget} confirms the same conclusion at approximately matched scale. AR remains clearly separated from the masked diffusion models on reverse KL, Wasserstein distance, energy distance, and TV distance. MSE again provides a much weaker separation, even though the generated samples differ substantially. The \(8\times8\) benchmark therefore exposes a support-recovery failure that is largely hidden by average grid-wise squared error.

\subsection{Investigating SEDD failure modes}
\label{subsec:sedd-failure-modes}

\begin{figure}[!htbp]
    \centering
    \includegraphics[width=0.72\linewidth]{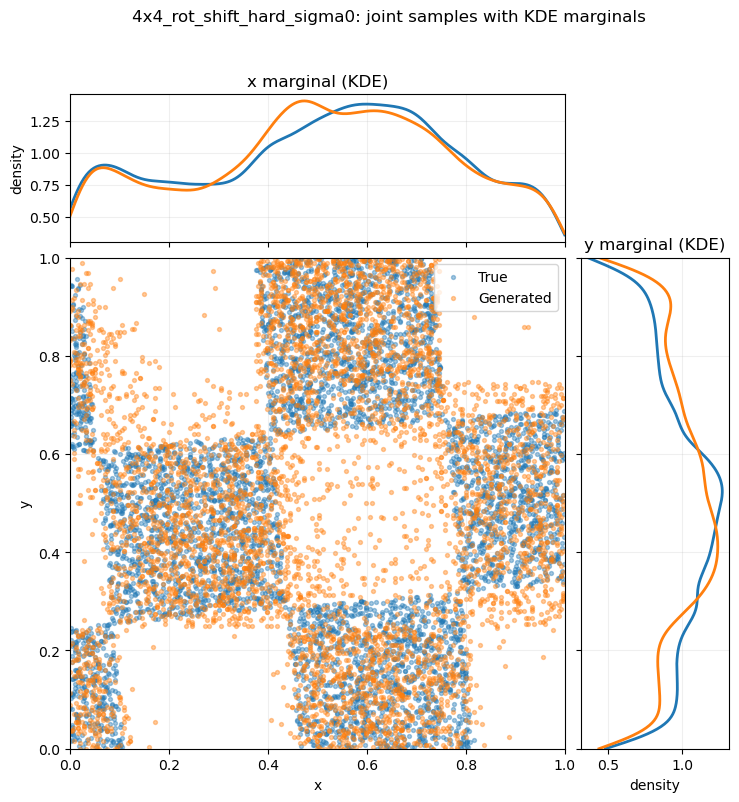}
    \caption{
    Joint samples and KDE marginals for SEDD on the rotated and shifted \(4\times4\) checkerboard. 
    Although the generated \(x\)- and \(y\)-marginals are close to the true marginals, the joint samples leak into invalid regions, suggesting that SEDD captures marginal structure more easily than the full two-dimensional dependency structure.
    }
    \label{fig:sedd-marginals}
\end{figure}

The checkerboard results suggest that SEDD fails in a more specific way than simple underfitting. In both the rotated \(4\times4\) setting and the harder \(8\times8\) setting, SEDD often captures coarse one-dimensional structure while failing to recover the full joint support. Figure~\ref{fig:sedd-marginals} illustrates this on the rotated and shifted \(4\times4\) checkerboard. The generated samples broadly match the true \(x\)- and \(y\)-marginals, but the joint distribution is blurred and places substantial mass in invalid regions. This supports the hypothesis that SEDD learns marginal statistics more easily than the higher-order dependencies needed to reconstruct the rotated checkerboard geometry.

\begin{figure}[!htbp]
    \centering
    \includegraphics[width=0.82\linewidth]{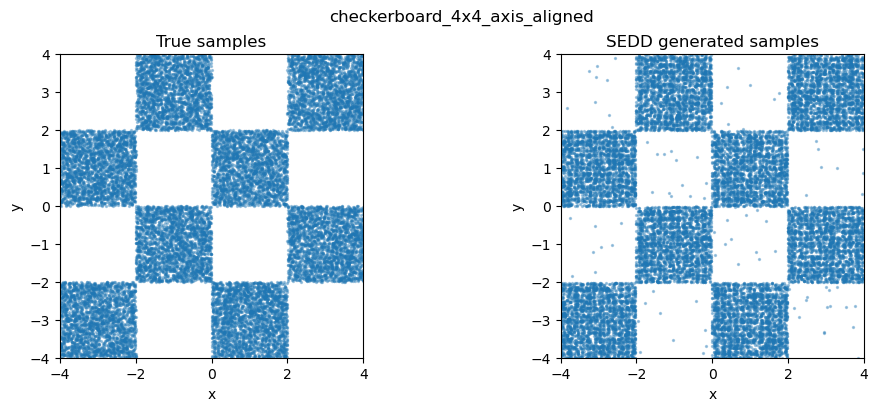}
    \caption{
    SEDD samples on the axis-aligned \(4\times4\) checkerboard. 
    In this symmetric setting, SEDD recovers the checkerboard support substantially better than in the rotated and shifted setting.
    }
    \label{fig:sedd-axis-four}
\end{figure}

\begin{figure}[!htbp]
    \centering
    \includegraphics[width=0.82\linewidth]{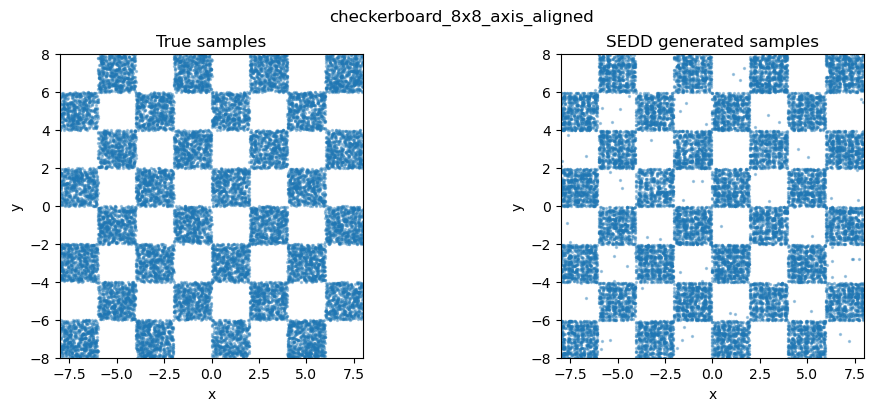}
    \caption{
    SEDD samples on the axis-aligned \(8\times8\) checkerboard. 
    Despite the larger number of cells, the regular axis-aligned structure remains comparatively easy for SEDD to capture.
    }
    \label{fig:sedd-axis-eight}
\end{figure}

This interpretation is consistent with the easier axis-aligned checkerboards. On the axis-aligned \(4\times4\) and \(8\times8\) settings, shown in Figures~\ref{fig:sedd-axis-four} and~\ref{fig:sedd-axis-eight}, SEDD recovers the support much more cleanly. These distributions are symmetric and aligned with the coordinate axes, so low-order marginal structure provides a stronger cue for the joint support. In the rotated setting, that shortcut no longer works: the support depends on coordinated changes across both coordinates and across multiple bits.


\begin{figure}[!htbp]
    \centering
    \includegraphics[width=0.82\linewidth]{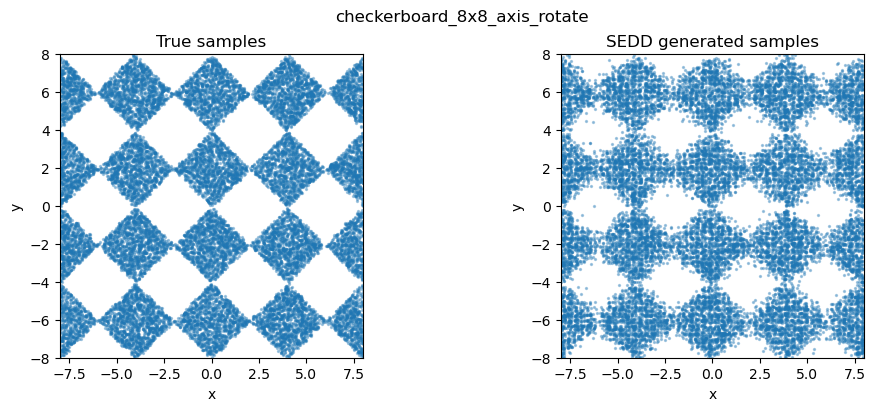}
    \caption{
    SEDD samples on the rotated \(8\times8\) checkerboard. 
    Compared with the axis-aligned case, rotation makes the support much harder to recover, producing diffuse samples and visible off-support mass.
    }
    \label{fig:sedd-rotated-eight}
\end{figure}

By contrast, the rotated \(8\times8\) checkerboard in Figure~\ref{fig:sedd-rotated-eight} is substantially harder. Rotation breaks the alignment between checkerboard support and coordinate-wise quantization. Matching the target therefore requires stronger joint dependencies across bits and coordinates. SEDD produces diffuse samples, rounds off sharp boundaries, and places mass between valid regions, matching its poor KL-based performance.

A plausible explanation is that SEDD is particularly sensitive to this benchmark because its score-ratio parameterization and sampling dynamics must assemble coherent samples through local discrete updates. This can work well for smoother targets, but the rotated checkerboard requires precise global coordination among many bits. Simpler training does not necessarily make the induced generation problem easier: the model still has to combine local token-level information into a globally valid two-dimensional support. In our experiments, SEDD appears to learn the easier marginal aspects of the distribution first but struggles to combine them into the correct rotated geometry.

\FloatBarrier

\begin{table*}[!t]
    \centering
    \caption{
    GMM results under standard binary and Gray-code encodings. Values are mean $\pm$ SEM over 10 independent seeds (0–9).
    }
    \label{tab:gmm_gray_extended}
    \small
    \setlength{\tabcolsep}{5pt}
    \resizebox{\textwidth}{!}{%
    \begin{tabular}{l r c c c c c c}
        \toprule
        & & \multicolumn{3}{c}{\textbf{Standard Binary}} & \multicolumn{3}{c}{\textbf{Gray Code}} \\
        \cmidrule(lr){3-5}
        \cmidrule(lr){6-8}
        \textbf{Model} & \textbf{Params} & \textbf{Forward KL} & \textbf{Reverse KL} & \textbf{MSE ($\times 10^{-9}$)} & \textbf{Forward KL} & \textbf{Reverse KL} & \textbf{MSE ($\times 10^{-9}$)} \\
        \midrule
        AR    & 6,322,435 & $10.21 \pm 0.17$ & $1.10 \pm 0.03$ & $7.52 \pm 0.08$ & $10.18 \pm 0.16$ & $1.09 \pm 0.02$ & $7.49 \pm 0.05$ \\
        MD4   & 4,739,842 & $10.58 \pm 0.17$ & $1.31 \pm 0.03$ & $7.40 \pm 0.06$ & $10.57 \pm 0.18$ & $1.28 \pm 0.03$ & $7.65 \pm 0.09$ \\
        MDLM  & 4,739,842 & $11.00 \pm 0.15$ & $1.62 \pm 0.02$ & $7.32 \pm 0.05$ & $10.98 \pm 0.17$ & $1.57 \pm 0.03$ & $7.43 \pm 0.07$ \\
        ReMDM & 4,739,842 & $11.29 \pm 0.15$ & $1.88 \pm 0.03$ & $7.43 \pm 0.07$ & $11.01 \pm 0.18$ & $1.67 \pm 0.05$ & $7.37 \pm 0.07$ \\
        SEDD  & 4,875,779 & $11.10 \pm 0.26$ & $1.56 \pm 0.11$ & $7.52 \pm 0.07$ & $11.23 \pm 0.21$ & $1.63 \pm 0.09$ & $7.66 \pm 0.06$ \\
        \bottomrule
    \end{tabular}
    }
\end{table*}

\begin{table*}[!t]
    \centering
    \caption{
    $4\times4$ checkerboard results under standard binary and Gray-code
    encodings. Values are mean $\pm$ SEM over 10 independent seeds (0–9).
    }
    \label{tab:checkerboard_gray_extended}
    \small
    \setlength{\tabcolsep}{5pt}
    \resizebox{\textwidth}{!}{%
    \begin{tabular}{l r c c c c c c}
        \toprule
        & & \multicolumn{3}{c}{\textbf{Standard Binary}} & \multicolumn{3}{c}{\textbf{Gray Code}} \\
        \cmidrule(lr){3-5}
        \cmidrule(lr){6-8}
        \textbf{Model} & \textbf{Params} & \textbf{Forward KL} & \textbf{Reverse KL} & \textbf{MSE ($\times 10^{-9}$)} & \textbf{Forward KL} & \textbf{Reverse KL} & \textbf{MSE ($\times 10^{-9}$)} \\
        \midrule
        AR    & 6,322,435 & $14.22 \pm 0.05$ & $1.82 \pm 0.02$ & $6.63 \pm 0.02$ & $14.22 \pm 0.05$ & $1.83 \pm 0.02$ & $6.63 \pm 0.02$ \\
        MD4   & 4,739,842 & $14.44 \pm 0.05$ & $2.85 \pm 0.08$ & $6.68 \pm 0.03$ & $14.42 \pm 0.04$ & $2.79 \pm 0.08$ & $6.66 \pm 0.04$ \\
        MDLM  & 4,739,842 & $14.55 \pm 0.04$ & $3.44 \pm 0.09$ & $6.69 \pm 0.02$ & $14.51 \pm 0.04$ & $3.27 \pm 0.12$ & $6.67 \pm 0.03$ \\
        ReMDM & 4,739,842 & $14.57 \pm 0.04$ & $3.57 \pm 0.06$ & $6.65 \pm 0.02$ & $14.54 \pm 0.04$ & $3.44 \pm 0.08$ & $6.66 \pm 0.02$ \\
        SEDD  & 4,875,779 & $14.84 \pm 0.16$ & $5.04 \pm 0.82$ & $6.73 \pm 0.05$ & $14.65 \pm 0.06$ & $4.01 \pm 0.27$ & $6.69 \pm 0.03$ \\
        \bottomrule
    \end{tabular}
    }
\end{table*}

\FloatBarrier

\subsection{Gray-code encoding}
\label{app:gray-code-encoding}
We additionally include Gray-code encoding in the benchmark to assess
whether the observed trends depend on the choice of discrete
representation. For this extended comparison, we report the largest model configuration (model dimension $=256$, layers $=6$) over 10 independent random seeds (0--9), with values reported as mean $\pm$ SEM. Overall, the relative
ranking of the methods is highly robust to the encoding choice, although
binary encoding tends to be slightly more difficult for some methods in
the checkerboard setting, most notably SEDD.

Table~\ref{tab:gmm_gray_extended} extends Figure~\ref{fig:gmm-kl-mse-scaling} by comparing standard
binary and Gray-code encodings for the GMM experiment. The results are
nearly unchanged across encodings, indicating that the scaling trends
observed in Figure~\ref{fig:gmm-kl-mse-scaling} are not specific to standard binary encoding.

Table~\ref{tab:checkerboard_gray_extended} extends Figure~\ref{fig:checkerboard-scaling} with the
same binary-versus-Gray-code comparison for the $4\times4$ checkerboard
experiment. The overall ranking is again preserved. Gray-code encoding
slightly improves several metrics for the masked diffusion models, with
the largest difference observed for SEDD, but does not alter the main
qualitative conclusions.

\section{Hyperparameters and Reproducibility Details}
\label{app:hyperparameters}

For the GMM experiments, all models were trained for 50,000 steps with a
batch size of 64 using AdamW (learning rate $1\times10^{-4}$,
$\beta=(0.9, 0.999)$, and no weight decay). We considered the following
11 combinations of model dimension and number of layers:
\[
(16,2),\ (16,4),\ (16,6),\ (32,4),\ (32,6),\ (64,4),\ (64,6),
\ (128,4),\ (128,6),\ (256,4),\ (256,6).
\]
This resulted in 11 hyperparameter configurations for each model and
random seed. The two-layer configurations with model dimensions of 32,
64, 128, and 256 were intentionally excluded to avoid highly imbalanced
width-to-depth architectures; a two-layer architecture was retained only
for a model dimension of 16.

For the checkerboard experiments, all models were trained for 40,000
steps with a batch size of 256 using AdamW (learning rate
$1\times10^{-4}$, $\beta=(0.9, 0.999)$, and weight decay $0.03$).
We used 4,000 linear warm-up steps with an initial learning-rate factor
of $0.1$ and applied gradient clipping with a maximum norm of $1.0$.
We considered two numbers of layers, $\{4,6\}$, and three model
dimensions, $\{64,128,256\}$, resulting in six hyperparameter
configurations for each model and random seed.

For both experiments, we evaluated each trained model using 20,000
generated samples. Each run was conducted on a single NVIDIA V100 32GB
GPU, taking approximately 20 minutes for the GMM experiments and one hour
for the checkerboard experiments. Unless otherwise specified, each experiment was repeated with 10 independent random seeds (0--9), and all uncertainties shown in shaded regions, error bars, and tables indicate $\pm 1$ SEM.


\FloatBarrier
\clearpage

\end{document}